\pdfoutput=1

\documentclass[11pt]{article}

\usepackage{acl}

\usepackage{times}
\usepackage{latexsym}

\usepackage[T1]{fontenc}

\usepackage[utf8]{inputenc}

\usepackage{microtype}

\usepackage{inconsolata}

\usepackage{graphicx}

\usepackage{xcolor}
\usepackage{colortbl}
\usepackage{placeins}
\usepackage{hyperref}
\usepackage{url}
\usepackage{amsmath,amssymb}
\usepackage{makecell}
\usepackage{microtype}
\usepackage{multirow}
\usepackage{subfigure}
\usepackage{arydshln}
\usepackage{color}
\usepackage{graphicx}
\usepackage{enumitem}
\usepackage{bbding}
\usepackage{tikz}
\usepackage{pgfplots}

\usepackage{booktabs}
\usepackage{pgfplots}
\usetikzlibrary{shapes}
\definecolor{redlight}{RGB}{255, 204, 204}  
\definecolor{redmedium}{RGB}{255, 153, 153} 
\definecolor{red}{RGB}{238,201,0}           
\definecolor{orange}{RGB}{255,99,71}           
\definecolor{plum}{RGB}{238,122,233}           
\definecolor{lsblue}{RGB}{176,196,222}           

\usepackage{inconsolata}

\author{Zhen Zhang$^{1}$, Xinyu Wang$^{2}$\thanks{\hspace{0.3em} Corresponding author}, Yong Jiang$^{2*}$, Zile Qiao$^{2}$, Zhuo Chen$^{3}$, Guangyu Li$^{3}$, \\ {\bf Feiteng Mu, Mengting Hu$^{1*}$, Pengjun Xie$^{2}$, Fei Huang$^{2}$} \\
        $^{1}$College of Software, Nankai University \\
        $^{2}$Tongyi Lab, Alibaba Group \\
        $^{3}$School of Information Science and Technology, ShanghaiTech University \\
        $^{3}$Shanghai Engineering Research Center of Intelligent Vision and Imaging \\
        \texttt{zhzhen23@gmail.com} \\
}

\title{KBM: Delineating Knowledge Boundary for Adaptive Retrieval in \\ Large Language Models}

\begin{document}
\maketitle

\begin{abstract}
Large Language Models (LLMs) often struggle with dynamically changing knowledge and handling unknown static information. Retrieval-Augmented Generation (RAG) is employed to tackle these challenges and has a significant impact on improving LLM performance.
In fact, we find that not all questions need to trigger RAG. By retrieving parts of knowledge unknown to the LLM and allowing the LLM to answer the rest, we can effectively reduce both time and computational costs.
In our work, we propose a \textbf{K}nowledge \textbf{B}oundary \textbf{M}odel (KBM) to express the known/unknown of a given question, and to determine whether a RAG needs to be triggered. 
Experiments conducted on 11 English and Chinese datasets illustrate that the KBM effectively delineates the knowledge boundary, significantly decreasing the proportion of retrievals required for optimal end-to-end performance. Furthermore, we evaluate the effectiveness of KBM in three complex scenarios: dynamic knowledge, long-tail static knowledge, and multi-hop problems, as well as its functionality as an external LLM plug-in.

\end{abstract}

\section{Introduction}\label{introduction}
As Large Language Models (LLMs) evolve, their real-world applications expand, yet they often struggle with dynamically changing and unknown static knowledge, leading to inaccuracies or \textit{hallucinations} \citep{rawte-etal-2023-troubling}. Retrieval-Augmented Generation (RAG) effectively addresses challenges by retrieving relevant external information in real time, enhancing LLMs' accuracy. However, it also incurs costs, such as increased retrieval requests and longer response times, leading to the crucial question: When is retrieval truly necessary?
\begin{figure}
    \centering
    \includegraphics[scale=0.40]{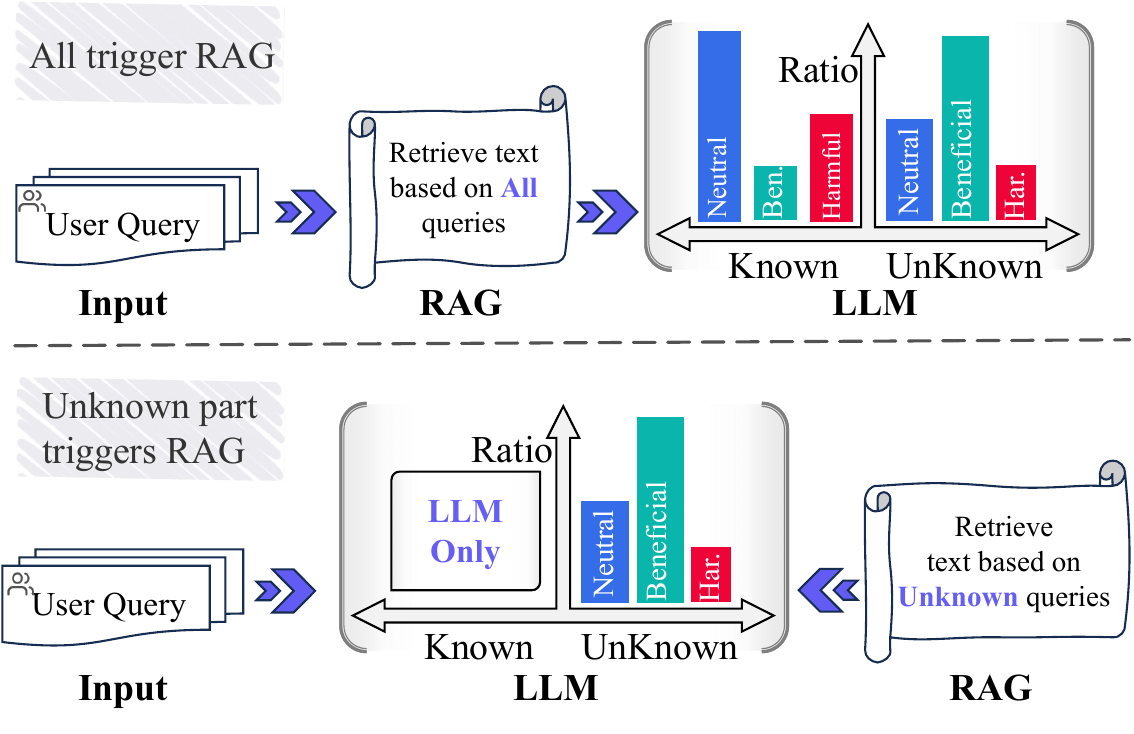}
    \caption{Illustration of the impact of different RAG triggering methods on LLM. \textcolor[rgb]{0.3294,0.6980,0.6667}{Beneficial}: RAG effectively solves this question. \textcolor[rgb]{0.2745,0.4314,0.8706}{Neutral}: RAG has no effect on the LLM. \textcolor[rgb]{0.8398,0.1484,0.2852}{Harmful}: RAG reduces the LLM's effectiveness.}
    \label{fig:fig_1}
\end{figure}

A natural criterion for determining the need for retrieval is the confidence level of the LLM. Specifically, the known knowledge of the model does not require retrieval, while uncertain parts can benefit from this process. As illustrated in Figure \ref{fig:fig_1} (top), using RAG for all questions (All RAG) can result in an overall impact that is neutral or even harmful, as the increased retrieval calls may not justify the benefits.
In contrast, Figure \ref{fig:fig_1} (bottom) demonstrates that when focusing solely on the uncertain part through RAG, the benefits are substantial, while reducing the frequency of unnecessary retrieval calls. The key to implementing this solution is how to enable LLM express confidence. A simple and effective solution is to use prompts. Unfortunately, over-/under-confidence will affect the expression of confidence \cite{xiong2024can}.

Previous studies on whether RAG is required for LLM can be divided mainly into two approaches. The first focuses on the \textit{question} itself, with methods like Self-RAG \citep{asai2023self} instructing models such as GPT-4 \citep{achiam2023gpt} to assess whether retrieving external documents (e.g., Wikipedia) can produce better responses. Although this approach can identify questions that require real-time information, it remains model-agnostic and struggles to determine whether an LLM has mastered specific knowledge.
The second approach evaluates both \textit{questions} and \textit{model responses} to determine if an LLM can answer a question, generating data by sampling multiple model responses and using manual labels for evaluation \cite{ren2024investigatingfactualknowledgeboundary,yin2023largelanguagemodelsknow,kadavath2022languagemodelsmostlyknow}. However, this method is labor intensive and relies heavily on manual labeling, which can create biases and lead to increased training costs. Additionally, the LLM's "unknown" expression has not been integrated with RAG triggering to assess its impact on end-to-end performance and retrieval ratio.

In this work, we propose the Knowledge Boundary Model (KBM) and two solutions for SFT based on \textit{accuracy} and \textit{certainty} in generating soft labels. The first method evaluates sampled QA accuracy and sets a threshold to classify questions as known or unknown. The second approach is based on certainty, focusing on modeling the entropy values of multiple sampled responses to establish thresholds, distinguishing between known and unknown responses through consistency, without requiring a golden answer. By defining different accuracy and certainty thresholds, we can generate data with soft labels, which is then used to fine-tune the KBM to determine whether the LLM considers a question unknown, thus indicating if retrieval is necessary.

We evaluate 11 English and Chinese QA datasets, demonstrating that KBM effectively measures knowledge mastery in LLMs. On average in these datasets, the certainty-based approach also reduces the number of retrievals by 13. 5\% compared with ALL RAG, resulting in a slightly higher performance of 0.1\%. Similarly, the accuracy-based approach reduces retrievals by 32.5\% on average, with only a 0.39\% performance drop.
Further analysis confirms the effectiveness of KBM in open domains, including dynamic knowledge, long-tail static knowledge, and multi-hop scenarios.

Our contributions are summarized as follows:

\begin{itemize}
    \item To our knowledge, KBM is the first study to reduce the RAG trigger ratio while maintaining LLM performance, thereby enhancing RAG's efficiency in QA tasks and reducing costs.
    \item For the technique contribution, we propose two methods for generating soft labels based on accuracy and certainty, allowing LLMs to express "unknown" or "known".
    \item KBM is validated on 11 datasets, demonstrating comparable effectiveness with All RAG and a reduced retrieval ratio, and it performs well in three complex scenarios.
\end{itemize}

\section{Preliminaries: LLM Knowledge Boundary and RAG Analysis}\label{preliminaries}
This section examines the impact of RAG on LLM performance. We evaluate the knowledge boundaries of LLMs with different parameter sizes and analyze RAG's effects on QA tasks, categorizing questions into three types based on its influence on LLM responses. Lastly, we introduce sampling methods based on LLM accuracy and certainty to further simulate knowledge boundary.

\subsection{How does RAG Affect the Accuracy of LLM Response?}



We evaluate how RAG affects the performance of LLMs with different parameter sizes. Our results show that while LLMs vary in their QA abilities and knowledge limits, their use of retrieved information is fairly consistent. We use three configurations: LLM Only generates responses from the LLM alone; ALL RAG enhances Naive RAG by adding the top 10 blocks retrieved from Google as context; and MASK RAG substitutes the correct answers in RAG with MASK, providing this altered data as context for the LLM.

We focus on the Qwen1.5 models (4B, 7B, 14B, 32B) \cite{bai2023qwentechnicalreport} alongside the Qwen2.0 72B model \cite{yang2024qwen2technicalreport}, using evaluation datasets for short question answering and reading comprehension tasks, including WebQA \citep{chang2022webqamultihopmultimodalqa}, SogouQA\footnote{https://github.com/sherlcok314159/ChineseMRC-Data}, and SQuAD1.5-zh\footnote{https://github.com/pluto-junzeng/ChineseSquad}. Our results show that LLM performance improves with larger parameter sizes across all three datasets, leading to a gradual differentiation in their QA capabilities. Notably, with Naive RAG, all models, especially those with 14B-72B parameters, effectively utilize contextual information. RAG has a more significant impact on smaller LLMs, but the maximum performance gain is similar across different model sizes. For example, on the WebQA dataset, the accuracy difference between the 7B and 72B models using RAG is just 3.05\%, compared to 28.87\% without RAG. Interestingly, MASK RAG seems to reduce the benefits of RAG, potentially harming performance on simpler datasets, as noisy information negatively affects smaller models more. For detailed results, see Appendix \S\ref{appendix_A} Figure \ref{fig:fig_2}.

These findings suggest that different LLMs possess varying knowledge boundary in question answering and demonstrate distinct retrieval strategies. Although all models exhibit strong capabilities in leveraging context, they show varying degrees of resilience to noise interference.

\subsection{Three Impacts of RAG on LLM}\label{sec_three_class}
To accurately assess the impact of RAG on LLMs, we categorize its effects into three aspects:
\begin{itemize}[leftmargin=*]
\item Beneficial: RAG effectively solves this question.
\item Neutral: RAG has no effect on the LLM.
\item Harmful: RAG reduces the LLM's effectiveness.
\end{itemize}
To isolate the impact of other modules, we use a simplified RAG pipeline for analysis, retrieving the top 10 blocks from Google Open Search to input into the LLM. To improve robustness, we sample each question 30 times. 
Inspired by \citet{kadavath2022languagemodelsmostlyknow}, we generate $I=30$ answer samples at temperature $T=1$. 
For a given question $Q$, if the model samples 20 correct answers and 10 incorrect ones, we construct an average accuracy of knowledge for that question based on these samples, resulting in a single data point $(Q, M_{pred} = \frac{20}{30})$. Unlike the approach in \cite{kadavath2022languagemodelsmostlyknow}, which includes 20 copies of (Q, M=1) and 10 copies of (Q, M=0) our method accurately represents the model’s understanding and misconceptions while significantly reducing the size of the training and test datasets by a factor of $I$. Thus, we approximate the model's soft labels for knowledge using hard labels derived from a diverse set of QA data points.


However, this method becomes challenging in the absence of gold-standard answers. To address this issue, we simulate the effect based on the certainty of the generated responses. Specifically, we compute the entropy distribution of words and phrases from the 30 generated answers to establish the model's level of certainty. Let $k$ represent the number of distinct answer types, denoted as $K_1, K_2,...,K_k$. The probability of each answer type occurring is represented as $P_1, P_2,...,P_k$, satisfying the normalization condition: $\sum_{i=1}^{k}P_i = 1$. Using this probability distribution, we quantify the model's certainty through entropy. First, we define entropy $H$ as:
\begin{figure}[t]
    \centering
    \subfigure[Category distribution based on \textbf{Accuracy} interval.]{\includegraphics[width=0.9\hsize,
    height=0.6\hsize]{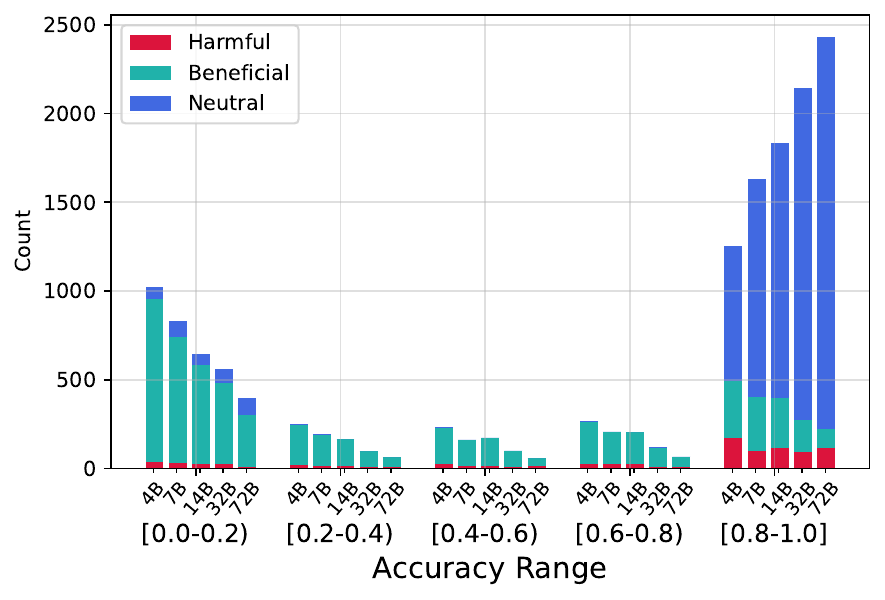}\label{fig:web_acc_norag}}{\hspace{-0.1cm}}
    \subfigure[Category distribution based on \textbf{Certainty} interval.]{\includegraphics[width=0.9\hsize,
    height=0.6\hsize]{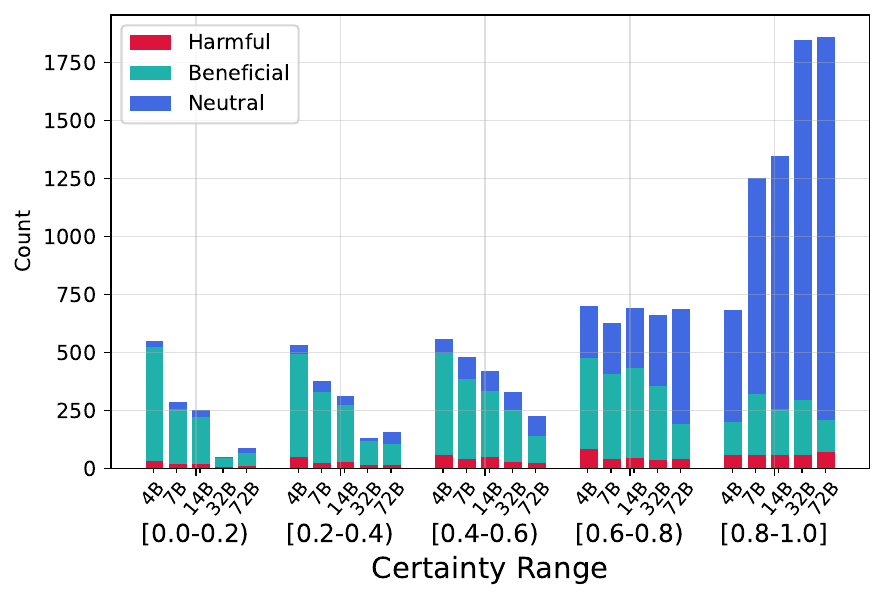}\label{fig:web_cer_norag}}{\hspace{-0.1cm}}
    \caption{Category distribution across various indicators: (a) Accuracy and (b) Certainty intervals.}
    \label{fig:index_distr}
\end{figure}

\begin{figure*}
    \centering
    \includegraphics[scale=0.43
]{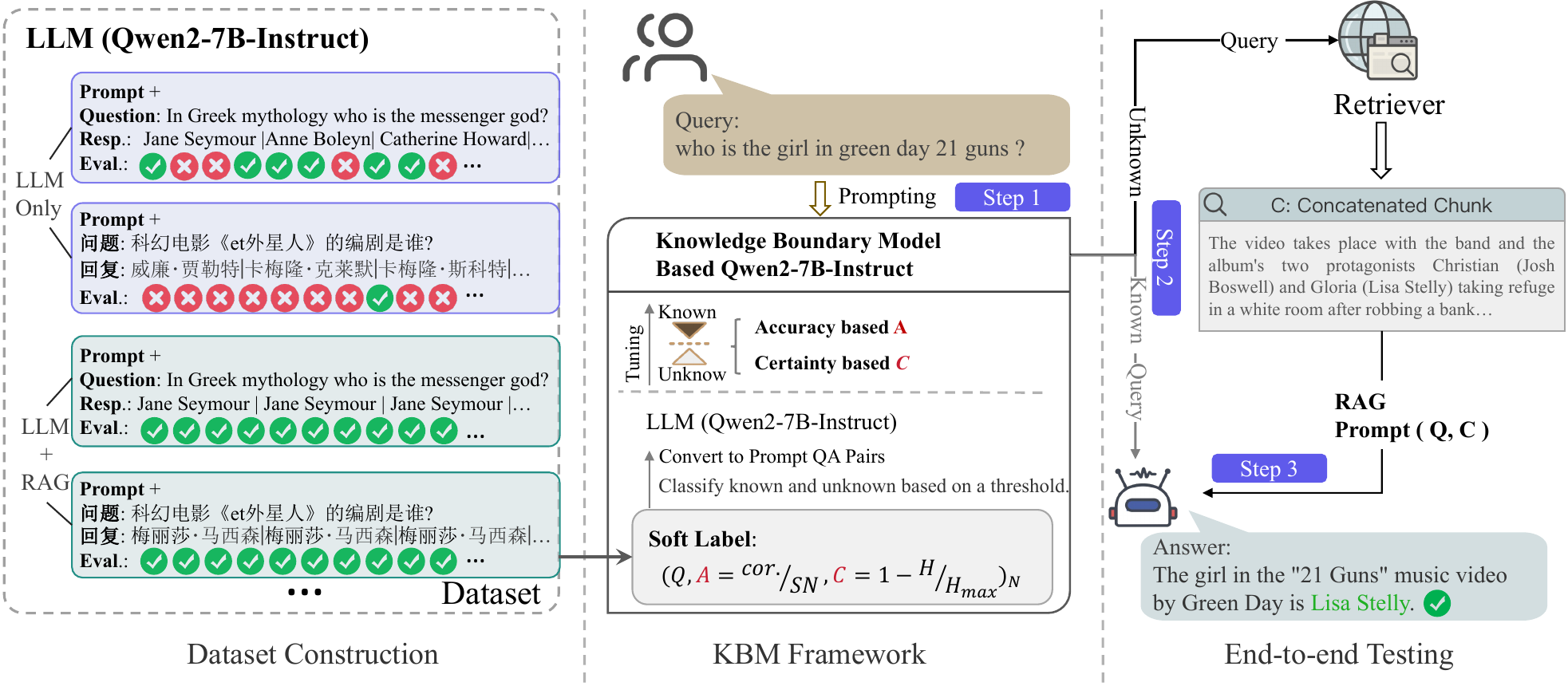}
    \caption{Illustration of the workflow for data generation, model training, and inference processes.}
    \label{fig:workflow}
\end{figure*}

\begin{equation}
H = -\sum_{i=1}^{k} P_i \log_2(P_i),
\end{equation}
and the maximum possible entropy $H_{max}=\log_2(k).$ The model's certainty $C(Q)$ is calculated by normalizing the entropy:
\begin{equation}
    C(Q) = 1 - \frac{H}{H_{{max}}},
\end{equation}
where ${H}/H_{max}\in[0, 1]$ is the normalized entropy.

We classify the effects of RAG based on the three categories and two indicators as follows: an effect is deemed \textbf{Beneficial} if the indicator increases after incorporating RAG compared to the LLM Only scenario; \textbf{Neutral} if the indicator remains unchanged and aligns with the value from the LLM Only condition; and \textbf{Harmful} if the indicator decreases following the addition of RAG, falling below the value observed in the LLM Only scenario. Using these categories, we analyze the class distribution of LLMs across accuracy intervals on the WebQA dataset, as shown in Figure \ref{fig:web_acc_norag}. The figure indicates that the [0-0.8) interval has the highest proportion of Beneficial cases, while Neutral and Harmful cases peak in the [0.8-1.0] range. This suggests a cost-benefit relationship: higher accuracy reflects greater confidence in answers, reducing the advantages of RAG, especially in the [0.8-1.0] range, where harmful cases also concentrate.
High certainty reflects high confidence, while low certainty indicates doubt. The certainty metric considers factors such as chance and model uncertainty \cite{hu2023uncertaintynaturallanguageprocessing}. We find a Pearson correlation coefficient of $0.64$ between accuracy and certainty, as shown in Figure \ref{fig:index_distr}. Certainty also has a similar category distribution to accuracy, indicating a positive correlation.

In the high accuracy range (above $0.8$), we first note that many neutral examples unnecessarily trigger RAG. Additionally, harmful examples can lead to a sharp accuracy drop, potentially from $1.0$ to $0.0$, while beneficial examples typically increase accuracy from 0.8 to a maximum of $1.0$.
This leads us to propose a threshold to differentiate between known and unknown data. Data below this threshold is marked as unknown and triggers RAG, while data above it is considered known, allowing the LLM to respond independently.

\section{Can LLM Express Know/Unknown?}\label{method}
In this section, we generate known and unknown soft labels based on accuracy and certainty for training the KBM, and we introduce the baselines and datasets used for evaluation.

\begin{table*}[htbp]\scriptsize
    \centering
    \begin{tabular}{p{0.6cm}lcccccccccccc} 
        \toprule
        & &\textbf{NQ} &\textbf{TriQA} &\textbf{OBQA}&\textbf{MMLU}&\textbf{SQD}&\textbf{MLEC}&\textbf{WebQA}&\textbf{SogQA}&\textbf{zh\_SQD}&\textbf{CMMLU}&\textbf{CSQA}&\textbf{Avg.}\\ 
        \midrule
        \multicolumn{10}{l}{\textbf{Naive} \textit{(Qwen2-7B-Instruct based + Google)}}\\
        \noalign{\vspace{0.1cm}}
        \rowcolor{lsblue!30}
        {LLM} & A/E & 35.58 & 50.79 & 82.72 & 66.45 & 34.02 & 74.25 & 70.06 & 52.53 & 22.72 & 74.23 & 24.13 & 53.41 \\ 
        \noalign{\vspace{0.01cm}}
        \rowcolor{redmedium!30}
        {RAG} & A/E & 61.63 & 84.60 & 81.48 & 70.80 & 89.83 & 68.40 & 92.54 & 83.01 & 83.66 & 77.05 & 77.80 & 79.16  \\ 
        \midrule
        \multicolumn{10}{l}{\textit{Qwen2-7B-Instruct based + Google}}\\
        \multirow{3}{*}{\textbf{Prompt}} & A/E & 59.92 & 79.08 & 81.96 & 67.30 & 78.55 & 72.80 & 89.75 & 78.13 &78.74 & 77.97 & 75.41 & 76.33 \\ 
        
        &Rnd. & 57.91 & 75.62 & 82.12 & 67.70 & 75.30 & 72.01 & 87.01 & 76.75 & 77.82 & 72.15 & 75.14 & 74.50\\  
        &Rat. & 86.0\% & 73.6\% & 27.8\% & 27.2\% & 75.6\% & 44.7\% & 77.5\% & 78.9\% & 90.3\% & 49.7\% & 94.1\% & 65.9\% \\  

        \midrule
        \multicolumn{10}{l}{\textbf{Self-RAG} \textit{(Llama-2 based + Google)}}\\

        \multirow{3}{*}{{\quad-\;7B}} & A/E & 31.07 & 69.82 & 74.80 & 47.93 & 62.92 & 29.48 & 52.96 & 39.83 & 29.47 & 33.94 & 18.03 & 44.57\\ 
        &Rnd. & 35.79 & 69.49 & 76.40 & 49.38 & 62.27 & 30.72 & 54.02 & 39.83 & 30.33 & 34.63 & 19.17 & 45.64\\  
        &Rat. & 22.7\% & 55.9\% & 10.8\% & 29.9\% & 77.0\% & 10.4\% & 28.3\% & 28.5\% & 37.0\% & 3.4\% & 23.2\% & 29.7\% \\  
        \midrule
        \multicolumn{10}{l}{\textbf{KBM} \textit{(Qwen2-7B-Instruct based  + Google)}}\\
        \multirow{3}{*}{\quad-\;{Acc.}} & A/E & 60.09 & 81.39 & \textbf{82.64} & 70.00 & 89.13 & \textbf{73.28} & 90.84 & 81.59 & 82.66 & \textbf{78.32} & 76.57 & 78.77\\ 
        &Rnd. & 51.20 & 74.42 & 81.92 & 68.59 & 83.14 & 72.66 & 81.19 & 74.17 & 81.29 & 77.73 & 74.87 & 74.65\\  
        &Rat. & 86.2\% & 70.3\% & 26.2\% & 70.0\% & 87.2\% & 30.2\% & 56.8\% & 75.2\% & 96.0\% & 50.0\% & 94.3\%  & 67.5\%\\  
        \noalign{\vspace{0.1cm}}
        \cdashline{1-14}
        \noalign{\vspace{0.1cm}} 
        \multirow{3}{*}{\quad-\;{Cer.}} & A/E & \textbf{61.27} & \textbf{83.32} & 82.32 & \textbf{70.45} & \textbf{89.38} & {71.69} & \textbf{92.12} & \textbf{82.89} & \textbf{83.45} & 77.12 & \textbf{77.38} & \textbf{79.22} \\ 
        &Rnd. & 55.16 & 80.66 & 82.88 & 69.14 & 88.61 & 69.46 & 90.71 & 81.06 & 83.31 & 77.06 & 76.85 & 77.72\\
        &Rat. & 97.5\% & 87.7\% & 57.8\% & 65.3\% & 97.3\% & 69.0\% & 89.7\% & 94.6\% & 99.3\% & 95.3\% & 97.8\% & 86.5\%\\  
        \bottomrule
    \end{tabular}
    \caption{Comparison of performance and trigger ratio (Rat. \%) metrics between KBM and baseline models. The English dataset utilizes EM as the performance metric, while the Chinese dataset employs accuracy as the evaluation metric. Each method is compared with the random RAG trigger score (Rnd.) at the same trigger ratio.}
    \label{tab:main_table}
\end{table*}

\subsection{Methods}

To construct training data for the KBM, we generate soft labels using the sampling method detailed in Section \S\ref{sec_three_class}. As shown in Figure \ref{fig:workflow}, each query is assessed under two configurations: LLM Only and LLM + RAG, providing scores for \textit{accuracy} and \textit{certainty}. A threshold of $\tau=0.9$ is set to classify data as \textit{known}  or \textit{unknown}. The data is then formatted into QA pairs for fine-tuning the Qwen2-7B model. For additional setup details, refer to Appendix \S\ref{appdix_A}.

During the inference phase of the KBM, the process unfolds in three steps: At \textbf{step 1}, after the user enters a query, the system packages this query through a prompt and sends it to the KBM to assess accuracy and certainty. At \textbf{step 2}, if the KBM judges the query as known, it forwards the original query directly to the answer generation model (Qwen2-7B instruct) to generate a response. If the query is judged as unknown, the process proceeds to step 3. At \textbf{step 3}, the system performs an open-domain retrieval using Google, based on the user's original query. The retrieved chunks of information are then spliced together. The original query and the connected chunks are combined using RAG prompt and subsequently fed into the answer generation model to produce the final response.

\subsection{Experiment Setting}

\textbf{Metrics}: For the English test data, we use Exact Match (EM) as the metric, while Accuracy is applied to the other datasets, represented as A/E. Additionally, we consider the retrieval ratio (Rat.) as a crucial metric.

\textbf{Baselines}: We establish a baseline using the following methods: (1) \textit{Prompt}: we use prompts from Qwen2-7B \cite{yang2024qwen2technicalreport}, Llama3-70B \cite{llama3modelcard}, Qwen2-72B \cite{yang2024qwen2technicalreport}, and GPT-4o \cite{openai2024gpt4technicalreport} to determine whether the current query requires retrieval. Each model is presented with the same retrieval query. (2) \textit{Random}: This method serves as a dynamic benchmark. By analyzing the ratio of each baseline method that triggers RAG, a query from the test set is randomly selected to initiate the RAG evaluation. This approach effectively assesses whether the model has genuinely identified the unknown query or merely "guessed" it.
Among related approaches, we select \textit{Self-RAG} (Llama-2 7B based) \citep{asai2023self} as baselines for our analysis. We use Google as the retriever for all methods, selecting the first 10 snippets as contextual information. The experiment focuses on end-to-end effectiveness, the reduction in retrieval ratio, and the differences between the baseline methods and the random approach.

\textbf{Datasets}: KBM's training and test data encompass a variety of task types, including short answer questions, multiple choice questions, reading comprehension, and multi-hop questions. We utilize TriviaQA (TriQA) \cite{joshi-etal-2017-triviaqa}, WebQA \cite{chang2022webqamultihopmultimodalqa}, and a combination of MMLU \cite{hendrycks2021measuring}, MLEC \cite{li-etal-2021-mlec}, and XieZhi \cite{gu2024xiezhi} training sets to train the KBM model. We categorize the test sets into two groups: \textbf{In-Domain}, which includes TriviaQA \cite{joshi-etal-2017-triviaqa}, WebQA \cite{chang2022webqamultihopmultimodalqa}, MLEC \citep{li-etal-2021-mlec}, and MMLU \cite{hendrycks2021measuring}, and \textbf{Out-Of-Domain}, which comprises NaturalQA(NQ) \cite{kwiatkowski-etal-2019-natural}, Open BookQA (OBQA) \cite{OpenBookQA2018}, SQuAD (SQD) \cite{rajpurkar2018knowdontknowunanswerable}, FreshQA \cite{vu2023freshllmsrefreshinglargelanguage}, SogouQA (SogQA) \footnote{https://github.com/sherlcok314159/ChineseMRC-Data}, the Chinese SQuAD (zh\_SQD) \footnote{https://github.com/pluto-junzeng/ChineseSquad}, CMMLU \cite{li2024cmmlumeasuringmassivemultitask}, and Chinese SimpleQA (CSQA) \cite{he2024chinesesimpleqachinesefactuality}. For details on the datasets, see the Appendix \S\ref{datasets}.




\begin{figure}[t]
    \centering
    \includegraphics[width=0.9\linewidth]{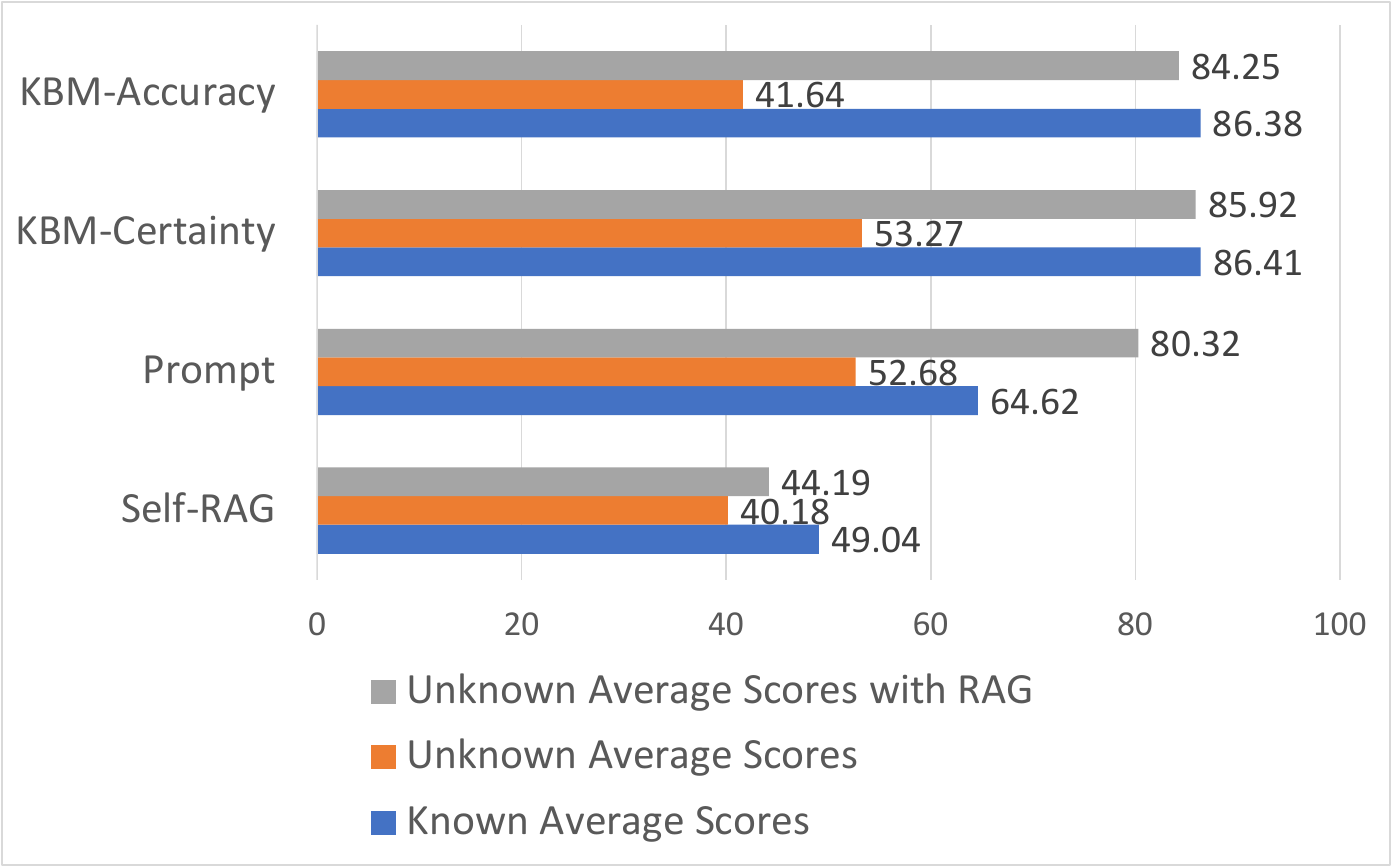} 
    \caption{The performance of LLM in the case of KBM model being judged as known/unknown (0-100). The higher the average score of known, the better.}
    \label{fig:unknown_known_res}
\end{figure}

\begin{figure}[t]
    \centering
    \includegraphics[width=0.9\linewidth]{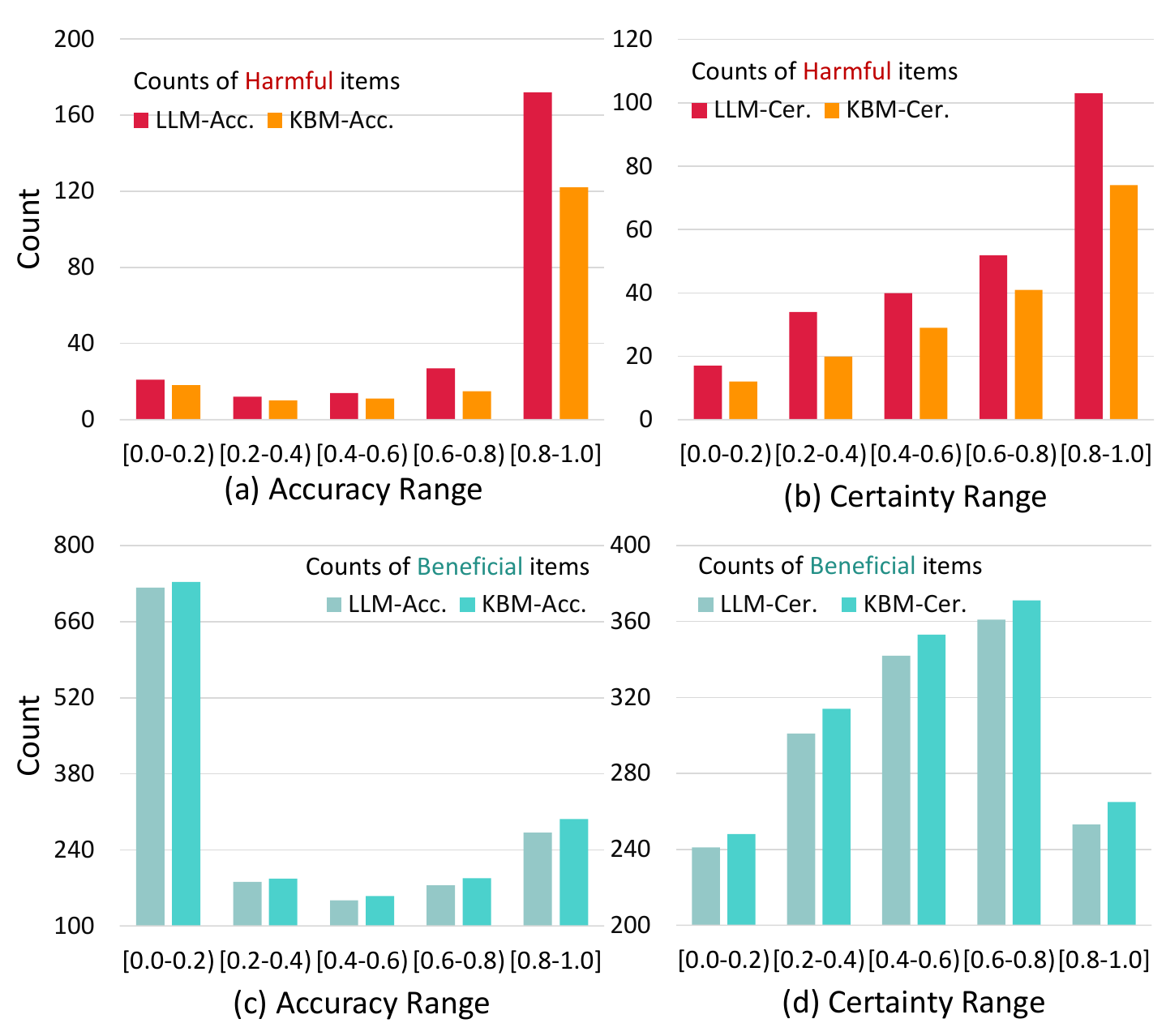} 
    \caption{Illustration of the KBM method's effects: the top shows a decrease in harmful counts, while the bottom indicates an increase in beneficial counts.}
    \label{fig:kbm_impact}
\end{figure}

\section{Experiments}\label{experiments}
In this section, we design extensive experiments to verify the effectiveness of KBM. 






\subsection{End-to-End Evaluation}
We evaluate the end-to-end performance of KBM across 11 test sets, showing that both accuracy and certainty reduce the RAG trigger ratio while enhancing the LLM's ability to answer questions. The results are presented in Table \ref{tab:main_table}. On average, the KBM accuracy-based method outperformed the LLM Only method by 25.4\%, narrowing the gap with All RAG to 0.4\%, while decreasing the RAG triggering ratio by 32.5\%, which is 4.1\% higher than the Random method, indicating a greater beneficial density. Similarly, the KBM certainty-based method improved performance by 25.8\% compared to the LLM Only method, achieving a performance that is 0.1\% higher than All RAG, and reduced the retrieval ratio by 13.5\%. Compared with the Prompt and Self-RAG methods, KBM effectively reduces the RAG triggering proportion while maintaining high performance.

Specifically, our analysis reveals that for short answer and QA test sets (e.g., NQ, TriviaQA, WebQA, SogouQA, and CSQA), both methods outperform the random baseline, with accuracy achieving particularly notable results. In reading comprehension tasks like SQA and zh\_SQD, where background information is essential for accurate answers, KBM exhibits a higher trigger ratio. All methods demonstrate strong performance and a high trigger ratio on FreshQA, indicating that KBM and LLM effectively capture queries with temporal features. However, in the context of multiple-choice questions found in CMMLU, MMLU, and OBQA, our method shows only marginal improvements over the random baseline. We guess that this limited enhancement arises from the challenges of sourcing relevant information for multiple-choice formats. Moreover, the diversity of training data formats, in relation to data type and task domain, is crucial for the effectiveness of KBM.


\textbf{Known/Unknown Expression Analysis}. KBM enables the model to effectively distinguish between known and unknown information, thereby optimizing performance and reducing costs. We analyze the mean scores across all test sets categorized as known or unknown by KBM, as illustrated in Figure \ref{fig:unknown_known_res}. 
We can see that without the assistance of RAG (represented by the orange bar), the LLMs struggle to handle unknown questions. In contrast, the two KBM methods achieve the most significant improvements compared to other approaches. Specifically, the prompt-based method shows `confusion' about the knowledge boundary because there is a notable gap between Known Aeverage Scores and Unknown Average Scores with RAG.
For improvements in accuracy and certainty at finer intervals, see the Appendix \S\ref{intervals_exp_Details}.

\begin{table}[t]\scriptsize
    \centering
    \begin{tabular}{lllllll}
        \toprule
        &\multicolumn{2}{c}{\textbf{fast-changing}} &\multicolumn{2}{c}{\textbf{slow-changing}}&\multicolumn{2}{c}{\textbf{never-changing}}\\
        &Ratio & Acc. & Ratio & Acc. & Ratio & Acc. \\
        \midrule
        \multicolumn{6}{l}{\textbf{Naive} \textit{(Qwen2-7B-Instruct based + Google)}}\\
        \noalign{\vspace{0.1cm}}
        \rowcolor{lsblue!30}
        {\textbf{LLM}}&0\%&29.7&0\% &29.8&0\%& 40.2 \\
        \noalign{\vspace{0.01cm}}
        \rowcolor{redmedium!30}
        \textbf{RAG}&100\%&50.4&100\% &60.2&100\%&66.6 \\
        \textbf{Prompt} & 93.8\% & 51.5 & 86.8\% & 57.8 & 81.8\% & 64.1 \\
        \noalign{\vspace{0.1cm}}
        \cdashline{1-7}
        \noalign{\vspace{0.1cm}}
        \textbf{Self-RAG}(7B) & 56.2\% & 37.4 & 56.3\% & 49.4& 55.1\% & 56.0 \\
        \textbf{KBM-Acc.} & 94.6\% & 51.6 & 92.3\% & 59.0 & 75.9\% & 62.7\\
        \textbf{KBM-Cer.} & 98.5\% & \textbf{52.0} & 95.6\% & \textbf{60.2} & 89.4\% & \textbf{64.6} \\
        \bottomrule
    \end{tabular}
    \caption{Results of triggering RAG with various levels of dynamic knowledge.}
    \label{tab:freq_qa}
\end{table}

\textbf{Impact Analysis of KBM on RAG}. In Section \S\ref{preliminaries} Figure \ref{fig:index_distr}, we discuss the three impacts of RAG on LLM. To investigate the changes in the number of harmful and beneficial questions, we employ KBM instead of RAG and find that KBM reduces the number of harmful questions while increasing the number of beneficial ones. Specifically, as shown in Figures \ref{fig:kbm_impact}(a) and \ref{fig:kbm_impact}(b), using WebQA as an example, we find that compared with RAG, the KBM-based approach decreases the number of harmful questions across all five intervals. This reduction is particularly evident in the high accuracy interval, indicating that we mitigate the harmful impact associated with RAG in areas close to what is known. Similarly, as illustrated in Figures \ref{fig:kbm_impact}(c) and \ref{fig:kbm_impact}(d), KBM increases the number of beneficial questions in each interval, showing that it enhances system performance by effectively triggering RAG.
Moreover, we conduct a comprehensive efficiency analysis, the results of which are presented in \S\ref{efficiency_analysis}.

\subsection{Analysis of Complex Scenarios}

A multitude of complex scenarios necessitate the frequent invocation of RAG. In light of this, we conducted tests and focused on the \textbf{ratio} of KBM in the following scenarios. 

\textbf{Dynamic Knowledge}.
We demonstrate that KBM effectively identifies questions with answers that change over time. Specifically, we classify the temporal changes in answers found in the FreshQA dataset into three distinct categories: fast-changing, slow-changing, and never-changing, based on the frequency of these changes.
In open domains, variations in answers often indicate the need for knowledge updates, necessitating the integration of external information into the LLM. As shown in Table \ref{tab:freq_qa}, the end-to-end performance without RAG for both fast-changing and slow-changing categories is suboptimal, highlighting a reliance on external knowledge. The high RAG trigger rates of KBM for these categories suggest that it effectively captures evolving answers. Conversely, the lower retrieval rate for the never-changing category implies that some knowledge is effectively embedded within the LLM. This finding underscores KBM's sensitivity in identifying questions with temporally correlated answers and highlights its role in enhancing dynamic knowledge adaptation.

\begin{table}[t]\small
    \centering
    \begin{tabular}{lllll}
        \toprule
        &\multicolumn{2}{c}{\textbf{en-FreshQA($\geq$2hop)}} & \multicolumn{2}{c}{\textbf{HotpotQA}}
        \\ 
        &Ratio&Acc.&Ratio&Acc.\\
        \midrule
        \noalign{\vspace{0.1cm}}
        \rowcolor{lsblue!30}
        \textbf{LLM}&0\% &26.1 &100\%& 30.5\\
        \noalign{\vspace{0.01cm}}
        \rowcolor{redmedium!30}
        \textbf{RAG}&100\% &48.7 &100\%& 51.7 \\
        \textbf{Prompt}&91.3\% &48.3 &93.7\%& 51.2\\
        \noalign{\vspace{0.02cm}}
        \cdashline{1-5}
        \noalign{\vspace{0.02cm}}
        \textbf{Self-RAG}(7B)&64.3\% &47.5 &65.0\%& 38.6\\
        \textbf{KBM-Acc.}&96.5\% &\textbf{48.9}& 91.2\% &50.9\\
        \textbf{KBM-Cer.}&99.1\% &48.7 & 94.3\% &\textbf{51.4}\\
        \bottomrule
    \end{tabular}
    \caption{Results of multi-hop triggering RAG.}
    \label{tab:second_table}
\end{table}

\textbf{Multi-Hop}.
A crucial aspect of our analysis is the ability of KBM to detect complex queries that necessitate multi-hop knowledge. Multi-hop questions comprise intricate knowledge components, requiring adjustments to the LLM. In these scenarios, KBM identifies the complexity of the queries and effectively employs RAG.
We tested queries involving two or more hops from the FreshQA($\geq$2-hop) and HotpotQA test sets, with results presented in Table  
\ref{tab:second_table}.
KBM demonstrates higher retrieval rates for multi-hop questions, indicating that it reveals more unknowns when addressing these complexities. As a result, the trigger ratio is higher than the average ratio for the task. While the overall end-to-end improvement is modest, this limitation stems from the need for further optimization of the RAG pipeline. We employ KBM to assess its ability to detect complex problems and trigger RAG, rather than to break them down and resolve them.

\begin{figure}[t!]
    \centering   
    \includegraphics[width=0.9\linewidth]{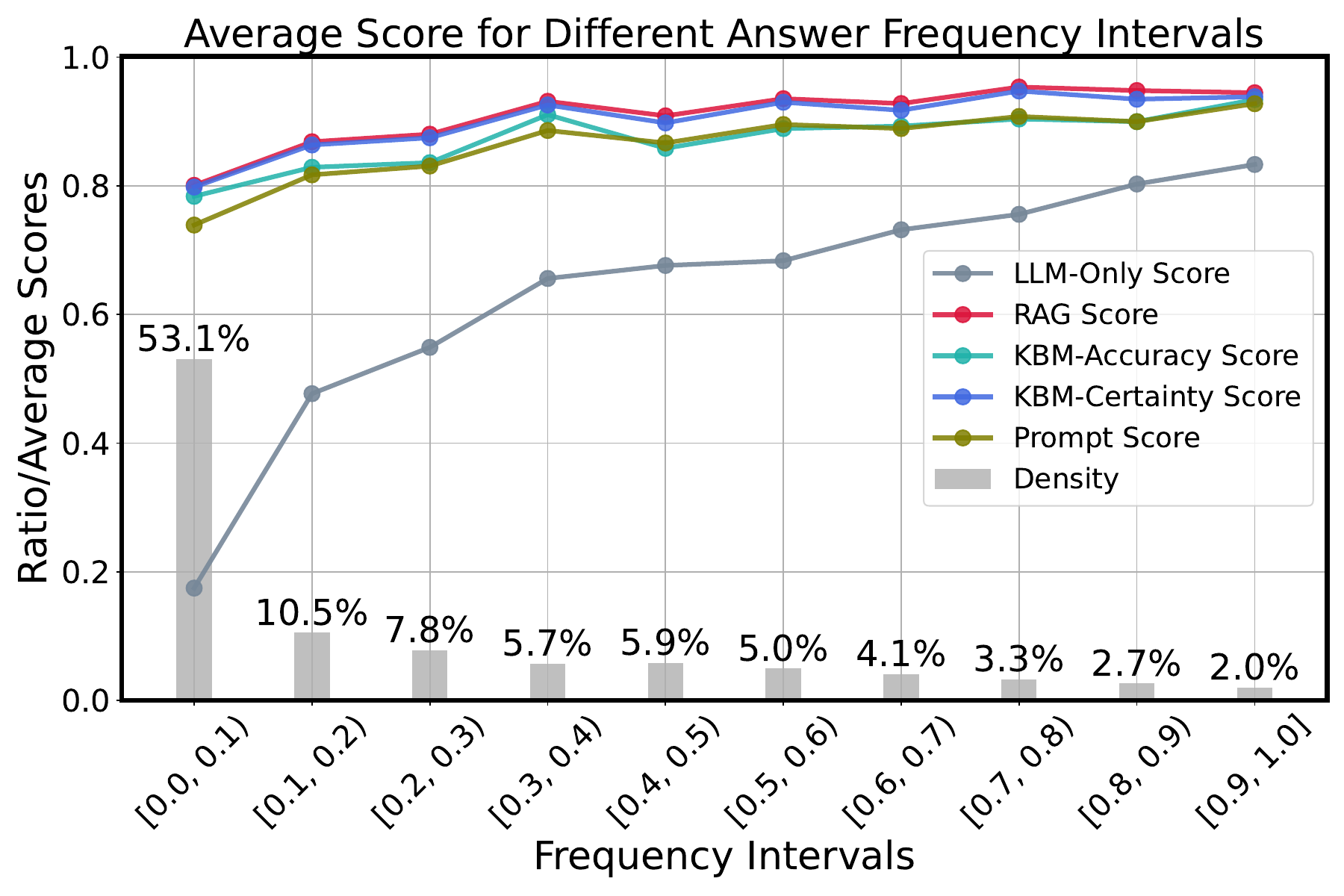} 
    \caption{Comparison of performance changes under different knowledge frequencies.}
    \label{fig:long_tail}
\end{figure}

\textbf{Long-tail Static Knowledge}. Long-tail knowledge has consistently posed challenges for the learning process of LLMs \cite{pmlr-v202-kandpal23a}.
We investigate the capacity of KBM to capture low-frequency long-tail knowledge across various question sets, confirming its effectiveness across all frequency ranges. Specifically, we combine test data from WebQA, SogQA, and zh\_SQD. Utilizing the gold answers from these datasets, we conduct vector retrieval within our Chinese database to differentiate knowledge based on its frequency. As illustrated in Figure \ref{fig:long_tail}, the LLM Only approach demonstrates reduced accuracy for low-frequency knowledge answers while performing better for high-frequency knowledge. However, integrating KBM with a prompt-based retrieval mechanism significantly enhances the model's performance for long-tail low-frequency knowledge. Notably, the certainty-based method yields the most substantial improvement, followed by the accuracy-based approach. These findings indicate that KBM effectively detects low-frequency long-tail knowledge and boosts overall performance by triggering RAG.
\begin{table}[t]\scriptsize
    \centering
    \begin{tabular}{p{0.8cm}p{0.2cm}cp{0.2cm}cp{0.2cm}cp{0.4cm}c} 
        \toprule
        & &\multicolumn{2}{c}{\textbf{\quad Gpt-4o}} &\multicolumn{2}{c}{\textbf{\; Qwen2-72B}} &\multicolumn{2}{c}{\textbf{Llama3-70B}}\\ 
        &&WebQA&NQ&WebQA&NQ&WebQA&NQ\\

        \midrule
        \noalign{\vspace{0.1cm}}
        \rowcolor{lsblue!30}
        {\textbf{LLM}} & A/E & 64.3&55.6&83.7&42.7&48.7&47.2\\ 
        \noalign{\vspace{0.01cm}}
        \rowcolor{redmedium!30}
        {\textbf{RAG}} & A/E & 90.0&60.1&91.9&59.4&90.9&60.2\\ 
        \noalign{\vspace{0.1cm}}
        \cdashline{1-8}
        \noalign{\vspace{0.1cm}}
        \multirow{3}{*}{\textbf{Prompt}} & A/E & 69.5&55.7&87.9&54.7&59.7&59.2\\ 
        &Rnd. & 60.8&56.6&85.6&52.6&55.3&58.6\\ 
        &Rat. & 14\%&18\%&22\%&51\%&16\%&47\%\\ 
        \noalign{\vspace{0.1cm}}
        \cdashline{1-8}
        \noalign{\vspace{0.1cm}}
        \multirow{3}{*}{\parbox{2cm}{ \textbf{Self-RAG}\\(7B)}} & A/E & 72.6&56.7&86.5&45.7&60.9&49.6\\ 
        &Rnd. & 71.6&57.1&86.0&42.7&48.7&47.2\\ 
        &Rat. & 28\%& 23\%& 28\%& 23\%& 28\%& 23\%\\ 
        \noalign{\vspace{0.1cm}}
        \cdashline{1-8}
        \noalign{\vspace{0.1cm}}
        \multirow{3}{*}{\textbf{KBM-Acc.}} & A/E & 85.7&60.6&91.4&58.5&77.9&59.6\\ 
        &Rnd. & 78.5&60.4&88.2&56.5&72.6&58.6\\ 
        &Rat. & 57\%& 86\%& 57\%& 86\%& 57\%& 86\%\\ 
        \noalign{\vspace{0.1cm}}
        \cdashline{1-8}
        \noalign{\vspace{0.1cm}}
        \multirow{3}{*}{\textbf{KBM-Cer.}} & A/E & \textbf{89.5}&\textbf{60.7}&\textbf{91.8}&\textbf{59.3}&\textbf{88.2}&\textbf{60.2}\\ 
        &Rnd. & 87.8&60.5&91.3&58.9&86.7&59.7\\ 
        &Rat. & 90\%& 98\%& 90\%& 98\%& 90\%& 98\%\\ 
        \bottomrule
    \end{tabular}
    \caption{End-to-end result with KBM as a plug-in for various LLMs' retrieval judgment modules.}
    \label{tab:plug_in}
\end{table}
\subsection{Use as a Plug-in}
To assess the effectiveness of the KBM model in triggering RAG, we apply it as a plug-in to GPT-4o, Llama3-70B, and Qwen2-72B. We evaluate these models using the WebQA and NQ test sets. The results are summarized in Table \ref{tab:plug_in}. Our findings indicate that KBM enhances the performance of these LLMs, although it does not achieve the comprehensive improvements provided by ALL RAG. This is consistent with our previous analysis showing that each LLM has a unique knowledge boundary, complicating the representation of knowledge boundary across all LLMs with a single model. The varying knowledge boundary of LLMs significantly influence the enhancement effects observed with knowledge boundary models (KBM). For example, the KBM-certainty judgment method shows that Qwen2-72B improves its performance from 83.7\% to 91.8\%, resulting in an increase of 8.1\%, while Llama3-70B experiences a much larger gain, rising from 48.7\% to 88.2\%, equating to a 39.5\% improvement. This illustrates that the initial knowledge capacity and boundary of an LLM can lead to divergent levels of enhancement, when interfacing with KBMs.

Additionally, RAG effectively enhances the response quality of LLMs, leading to a relatively stable upper limit for improvements across models. However, the performance gains differ due to the varying distributions of known and unknown knowledge in KBMs versus general models. For instance, although Llama3-70B demonstrates a notable retrieval rate of 90\%, its upper limit is lower than that of Qwen2-72B, which struggles to reach the ALL RAG score despite the same retrieval performance in the WebQA dataset. These trends are echoed in English QA tests, where significant performance improvements are observed when the English proficiency of the KBM ontology model inferior to that of the generative LLM. This suggests that aligning the knowledge boundary with the capabilities of the LLM is essential for optimizing performance.

\section{Related Work}\label{related_work}

\textbf{Large Language Model Knowledge Exploration}. The exploration of knowledge boundary in LLMs attracts significant attention. \citet{kadavath2022languagemodelsmostlyknow} examines the self-evaluation capabilities of LLMs, showing that larger models enhance their calibration by initially proposing answers and then evaluating their validity. 
\citet{ren2024investigatingfactualknowledgeboundary} studies LLMs’ perception of factual knowledge boundary and finds that they often display blind confidence in their abilities.
\citet{yin2023largelanguagemodelsknow} focuses on self-awareness, demonstrating that while LLMs can identify some unanswered questions, substantial discrepancies still exist, affecting their uncertainty detection. 
\citet{chen2024teachinglargelanguagemodels} introduces COKE, an unsupervised method for teaching models to articulate their knowledge limits through internal signals, yielding improved outcomes across various datasets. 
\citet{kang2024unfamiliarfinetuningexamplescontrol} points out that LLMs often default to examples in training data when facing unfamiliar queries. 
\citet{li2024knowunknownuncertaintysensitivemethod} explores hallucinations related to insufficient prompt context, showing that models frequently fail to recognize inadequate information.

\textbf{Retrieval-Augmented Generation}. RAG enhances LLMs by integrating retrieved text passages, significantly improving performance in knowledge-intensive tasks. A key focus is optimizing the timing and strategy of retrieval. \citet{asai2023self} introduce SELF-RAG, a method that trains LLMs to retrieve information, generate content, and evaluate their outputs using reflection tokens. This method enables the customization of model behavior, demonstrating significant performance improvements over standard RAG approaches. \citet{jeong2024adaptiveraglearningadaptretrievalaugmented} proposes Adaptive-RAG, which adjusts query handling based on complexity. \citet{wu2024clashevalquantifyingtugofwarllms} explores how LLMs process erroneous retrieved content. By creating a dataset to assess model responses to incorrect information, the study reveals insights into how models correct their outputs or may perpetuate errors. \citet{cuconasu2024power} conducts a comprehensive study on the retriever's function in RAG. 
\section{Conclusion}\label{conclusion}
In this paper, we propose KBM to address the limitations of LLMs in managing dynamic knowledge and unknown static information. KBM judges whether to trigger RAG by indicating the known and unknown of a question, preserving RAG's advantages while reducing computational costs.
Validated on 11 datasets, KBM matches the performance of All RAG while reducing the retrieval ratio, enhancing efficiency in QA tasks without compromising other instruction fine-tuning performance. We also confirm KBM's robustness in complex scenarios, such as dynamic and long-tail knowledge and multi-hop problems, highlighting its effectiveness as an external LLM plug-in. Our contributions include methods for generating soft labels based on accuracy and certainty, enabling LLMs to better express their knowledge status.

\section*{Limitations}\label{Limitations}
Our work is the first attempt to explore how to reduce the proportion of retrievals that trigger RAG while maintaining high performance. Despite the reduction in the proportion of triggers, our study has some limitations that may guide future efforts.

First, we propose a method based on accuracy and certainty sampling to generate soft labels for training KBM, demonstrating that LLMs learn to express unknowns in different tasks without requiring a large amount of data. However, in real-world scenarios, it is essential to include more diverse QA format data for better generalization, such as reasoning and content generation. Additionally, converting data from various question-answer formats into the format of QA pairs enhances the performance of KBM.

Secondly, while we enable LLMs to express unknowns and trigger RAG, there remains a relatively low probability that an unknown question posed by LLM can be resolved by RAG, particularly when considering post-retrieval operations such as changing the retriever, sorting, and denoising.

\bibliography{custom}

\appendix






\FloatBarrier

\begin{figure*}[t]
    \centering
    \includegraphics[scale=0.4]{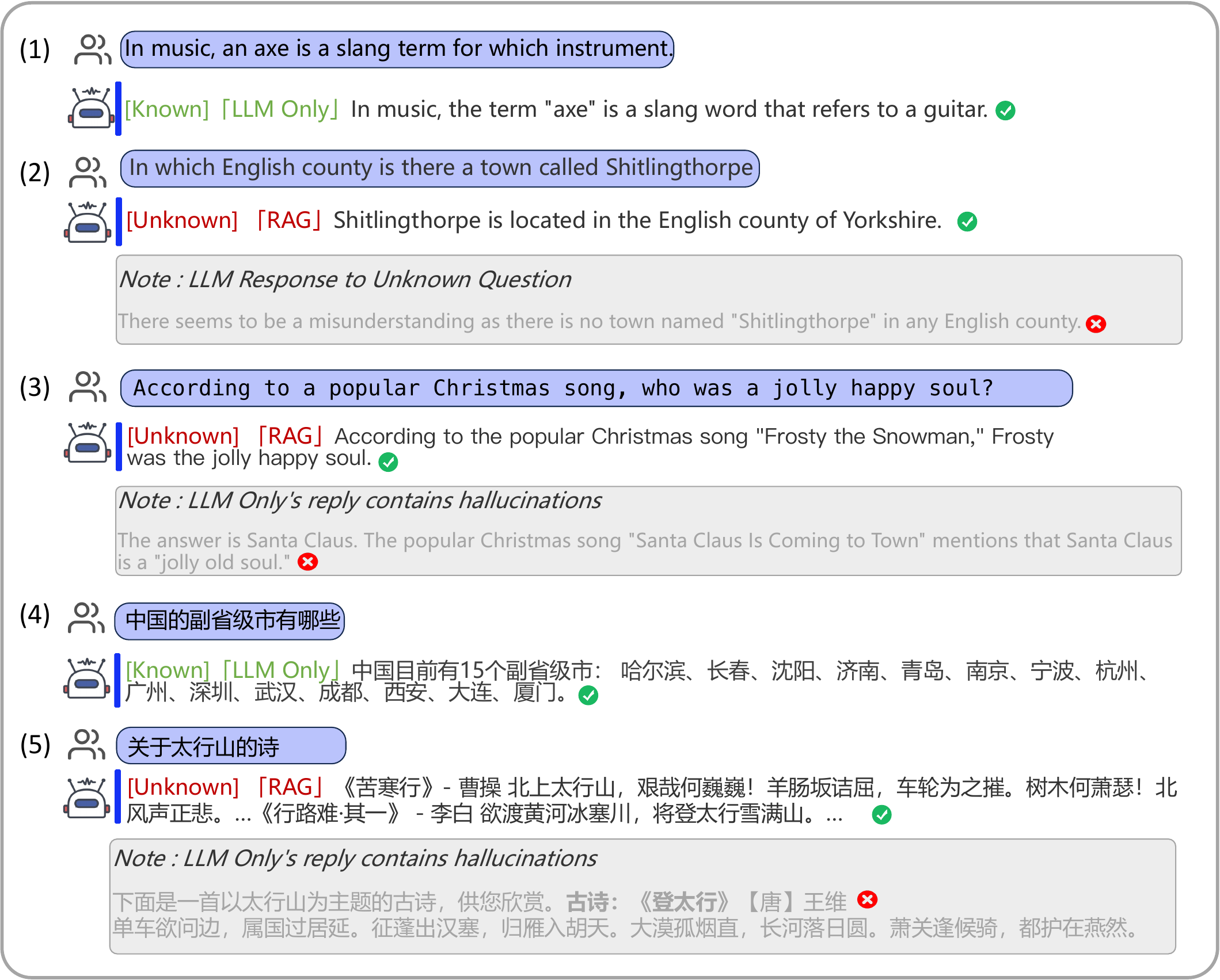}
    \caption{Case study of the KBM and effects of combining KBM with SFT data in a QA scenario.}
    \label{fig:case_study}
\end{figure*}
\FloatBarrier
\begin{figure}[t]
    \centering
    \hspace{-0.03\textwidth}
    \includegraphics[scale=0.36
]{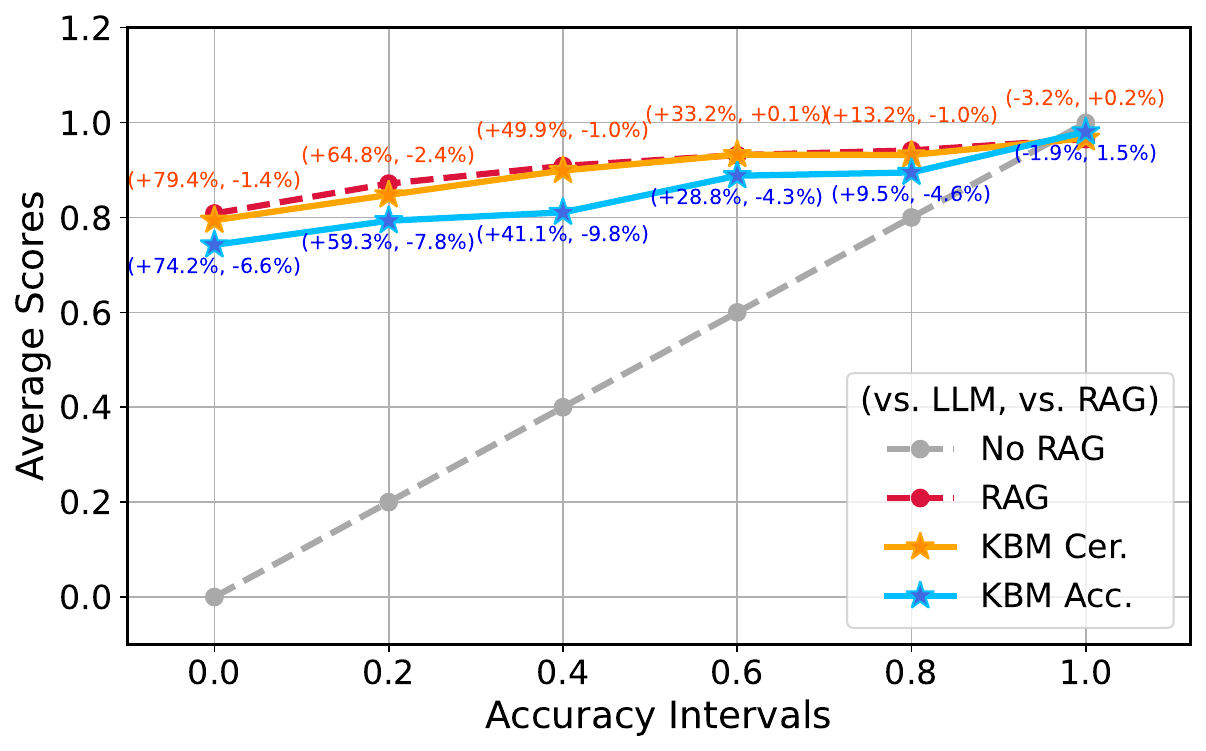}
    \caption{Illustration of the relative improvement of KBM for different accuracy ranges of LLM Only.}
    \label{fig:average_scores}
\end{figure}
\begin{figure}
    \centering
    \hspace*{-0.7cm}
    \includegraphics[scale=0.25]{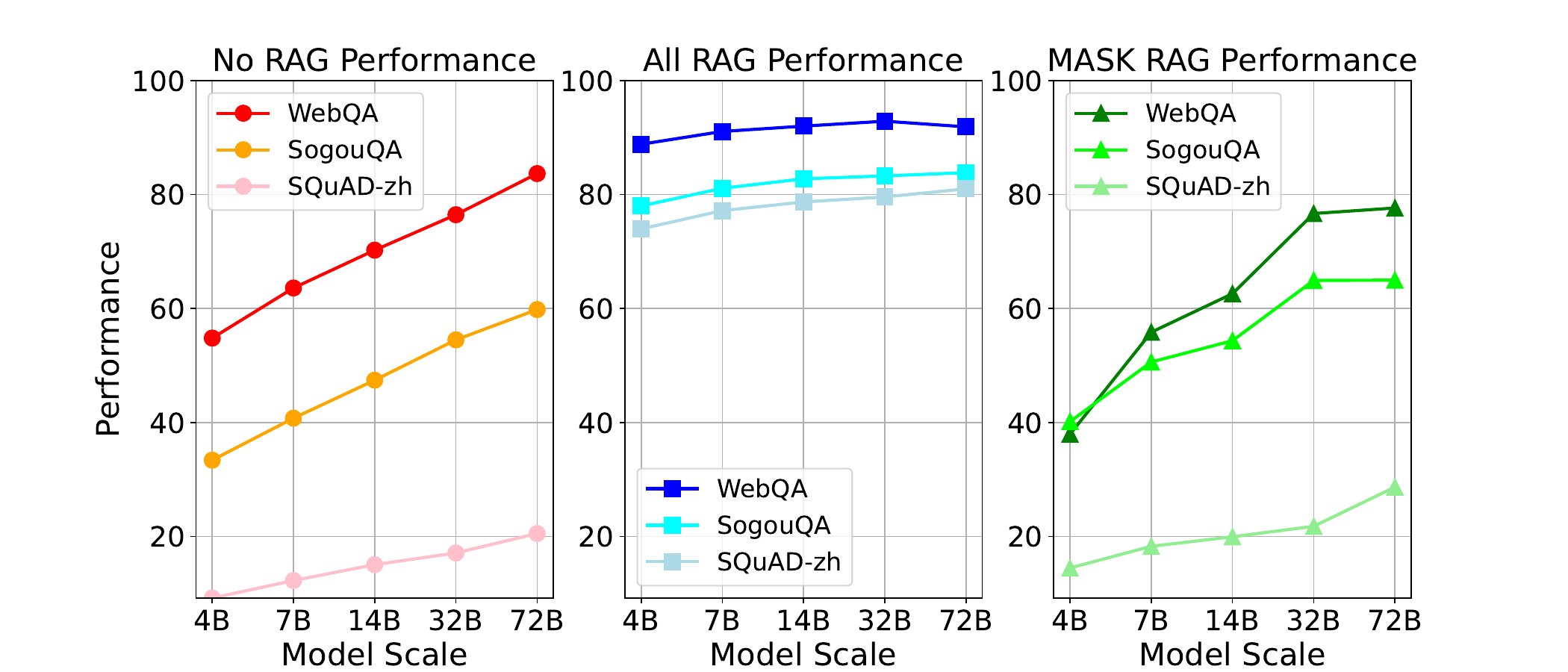}
    \caption{Illustration of the impact of RAG on LLM \\ performance.}
    \label{fig:fig_2}
\end{figure}
\section{Additional Experimental Analysis}\label{appendix_A}
\subsection{Analysis of KBM in Sub-intervals}\label{intervals_exp_Details}
We divided the accuracy interval into steps of 0.2 on the Chinese WebQA test set to investigate the improvements of KBM compared to RAG. As illustrated in Figure \ref{fig:average_scores}, our findings indicate that the KBM method, based on Accuracy and Certainty, demonstrates significant enhancements across several intervals, particularly in the lower accuracy ranges. For instance, when the LLM-only accuracy is at 0.0, KBM-Certainty shows an increase of 79.4\%, while KBM-Accuracy exhibits a rise of 74.2\%. Notably, when compared to All RAG, KBM achieves performance levels that are comparable to RAG in multiple intervals, especially within the Certainty-based approach.


\subsection{Case Study}\label{case_study_exp_Details}
As illustrated in Figure \ref{fig:case_study}, we showcase how KBM functions within a QA context. For questions (1) and (4), if KBM determines that the answer is Known, the LLM can provide the correct response directly, eliminating the need to search for external resources. In the case of question (2), which the LLM cannot answer or chooses not to address, KBM can preemptively classify it as unknown. This classification enables subsequent actions to penalize the search tool, prompting it to retrieve relevant textual materials that can then be provided to the LLM as context to aid in formulating an answer. For questions (3) and (5), direct use of the LLM to respond may lead to hallucinations, resulting in the output of incorrect information. In these instances, KBM can classify the question as unknown and activate RAG to produce a more reliable answer.

\begin{table}[htbp]\scriptsize
    \centering
    \begin{tabular}{lllll}
        \toprule
        &&Search API&Token&time\\
        &A/E&Call Rate&consumption&(ms)\\
        \midrule
        \noalign{\vspace{0.02cm}}
        \textbf{All RAG}&79.16 &100.00\% &593& 2498\\
        \textbf{KBM-Acc.}&78.77 &\textbf{67.50\%}& \textbf{421}&\textbf{2231}\\
        \textbf{KBM-Cer.}&\textbf{79.22} &86.50\% & 517 &2384\\
        \bottomrule
    \end{tabular}
    \caption{Efficiency analysis across 11 datasets.}
    \label{tab:eff_table}
\end{table}

\subsection{Efficiency analysis}\label{efficiency_analysis}
We evaluate our method across 11 datasets and compare it to the All RAG approach. The findings appear in Table \ref{tab:eff_table}. For KBM-Acc, the API calls reduce by \textbf{32.5\%}; tokens input to LLM decrease from 593 to 421; and the total computation time reduces from 2498 ms to 2231 ms. These results indicate that KBM-Acc offers significant advantages in both computational and invocation costs compared to All RAG. On the other hand, when comparing with All RAG, our KBM-Cer reduces all three metrics, namely API calls, token consumption, and time, to varying degrees. Importantly, KBM-Cer not only controls costs but also \textbf{surpasses All RAG} in overall performance metrics.

\section{Implementation Details}\label{appdix_A}

\subsection{Parameter Settings}\label{appdix_A1_para}
We employ four Nvidia A100 GPUs, each with 80GB of memory, to train our KBM models. To maintain high end-to-end performance, we set the threshold for both Accuracy and Certainty at 0.9. Each model undergoes training for three epochs, utilizing a batch size of four. The peak learning rate is set to 1e-5, with a warmup ratio of 2\% and cosine decay for the learning rate. To accommodate memory limitations, we restrict the maximum token length to 1580 for the 7B model and 1524 for the 13B model. For multi-GPU distributed training, we utilize Deepspeed stage 2 \citep{rajbhandari2020zero} while enabling Bfloat16 precision. Inference on the trained models is conducted using a single Nvidia Tesla V100 GPU with 32GB of memory.


\subsection{Datasets}\label{datasets}
\textbf{English Datasets}. NaturalQA \citep{kwiatkowski-etal-2019-natural}: This QA dataset, curated by Google, consists of real-world questions derived from natural retrieval queries. TriviaQA \citep{joshi-etal-2017-triviaqa}: This dataset is based on encyclopedic content and features complex questions and answers, primarily sourced from competitions and quizzes. MMLU \citep{hendrycks2021measuring}: This dataset comprises multiple-choice questions across various fields, assessing the model's knowledge mastery in academic and professional domains. OpenBookQA \citep{OpenBookQA2018}: Focusing on scientific inquiries, this dataset requires reasoning rooted in principles and common sense. en-SQuAD-en2.0 \citep{rajpurkar2018knowdontknowunanswerable}: This dataset features question-answer pairs and evaluates reading comprehension skills. FreshQA-en \citep{vu2023freshllmsrefreshinglargelanguage}: This dataset presents various question and answer types, offering a comprehensive assessment of QA capabilities. HotpotQA \citep{yang-etal-2018-hotpotqa}: This dataset consists of 113,000 Wikipedia based QA pairs that necessitate complex reasoning across multiple supporting documents and include sentence-level supporting facts.

\textbf{Chinese Datasets}. For the Chinese dataset, we utilize the following resources: WebQA \citep{chang2022webqamultihopmultimodalqa}: This open-domain QA dataset is collected via web crawlers, covering a wide array of topics and evaluating the performance of QA systems. SogouQA \footnote{https://github.com/sherlcok314159/ChineseMRC-Data}: Provided by Sogou, this dataset features user-generated questions and system-generated answers, assessing accuracy and robustness. MLEC \citep{li-etal-2021-mlec}: This dataset is designed to test the comprehension capabilities of models in various contexts. Xiezhi \citep{gu2024xiezhi}: A set of 249,587 Chinese/English questions covering 516 subjects for evaluating LLMs. SQuAD-zh \footnote{https://github.com/pluto-junzeng/ChineseSquad}: The Chinese version of the English SQuAD \citep{rajpurkar2018knowdontknowunanswerable} dataset serves to train and evaluate machine reading comprehension and QA systems. Chinese SimpleQA(CSQA) \citep{he2024chinesesimpleqachinesefactuality}: A benchmark for evaluating the factuality of LLMs in short Chinese QA across diverse topics.
CMMLU \citep{li2024cmmlumeasuringmassivemultitask}:This comprehensive benchmark assesses the knowledge and reasoning capabilities of language models across 67 topics, from basic to advanced.

\end{document}